%% file: main.tex
\documentclass[]{acmart} 

\usepackage{pgfplots}
\usepackage{pgfplotstable} 
\pgfplotsset{compat=1.18}
\usepackage{amsmath}

\usepackage{tikz,pgfplots,pgfplotstable}
\usepgflibrary{patterns} 
\usetikzlibrary{patterns}
\AtBeginDocument{%
  }

\copyrightyear{2025}
\acmYear{2025}
\setcopyright{rightsretained}
\acmConference[SCA '25]{SupercomputingAsia 2025}{March 10--13, 2025}{Singapore, Singapore}
\acmBooktitle{SupercomputingAsia 2025 (SCA '25), March 10--13, 2025, Singapore, Singapore}
\acmPrice{}
\acmDOI{10.1145/3718350.3718357}
\acmISBN{979-8-4007-1250-0/25/03}

\begin{document}
\title{Should AI Optimize Your Code? A Comparative Study of Classical Optimizing Compilers Versus Current Large Language Models}

\author{Miguel Romero Rosas}
\authornotemark[1]
\email{miguelro@udel.edu}
\orcid{0009-0003-7145-3429}
\author{Miguel Torres Sanchez}
\authornotemark[1]
\email{mgltorsa@udel.edu}
\orcid{0009-0003-6209-9675}
\affiliation{%
  \institution{University of Delaware}
  \city{Newark}
  \state{Delaware}
  \country{USA}
  \postcode{19713}
}

\author{Rudolf Eigenmann}
\email{eigenman@udel.edu}
\orcid{0000-0003-1651-827X}
\affiliation{%
  \institution{University of Delaware}
  \city{Newark}
  \state{Delaware}
  \country{USA}
  \postcode{19713}}

\renewcommand{\shortauthors}{Trovato et al.}


\begin{abstract}

Traditional optimizing compilers have played an important role in adapting to the growing complexity of modern software systems. The need for efficient parallel programming in current architectures requires strong optimization techniques. The beginning of Large Language Models (LLMs) raises intriguing questions about the potential of these AI approaches to revolutionize code optimization methodologies. This work aims to answer an essential question for the compiler community: "Can AI-driven models revolutionize the way we approach code optimization?". 

To address this question, we present a comparative analysis between three classical optimizing compilers and two recent large language models,  evaluating their respective abilities and limitations in optimizing code for maximum efficiency. In addition, we introduce a benchmark suite of challenging optimization patterns and an automatic mechanism for evaluating the performance and correctness of the code generated by LLMs. We used three different prompting strategies to evaluate the performance of the LLMs – Simple Instruction (IP), Detailed Instruction Prompting (DIP), and Chain of Thought (CoT). 

A key finding is that while LLMs have the potential to outperform current optimizing compilers, they often generate incorrect code on large code sizes, calling for automated verification methods. In addition, expressing a compiler strategy as part of the LLM’s prompt substantially improves its overall performance. Our evaluation across three benchmark suites shows CodeLlama-70B as the superior LLM, capable of achieving speedups of up to x1.75. Additionally, CETUS is the best among the current optimizing compilers, achieving a maximum speedup of 1.67x. We also found substantial differences among the three prompting strategies.

\end{abstract}



\footnote{Both authors contributed equally to this research. Accepted at SupercomputingAsia (SCA'25), March 10-13,2025}


\maketitle
\input{introduction}
\input{LLMsVsCompilers-limitations-strengths}
\input{EnvironmentEvaluation}

\input{LLMsVSCompilers}

\input{Related_Work}

\input{finding-and-conclusions}
\input{acknowdledgements}

\bibliographystyle{ACM-Reference-Format}
\bibliography{main}
\end{document}

%% file: introduction.tex
\section{Introduction}
\label{sec:introduction}
The need for developing and optimizing parallel programs is more crucial than ever~\cite{doi:10.1126/science.aam9744}. Optimizing compilers have long been one of the pillars of creating parallel code, evolving continuously to meet the growing complexity of modern programs and architectures. However, despite all advancements, current compilers still do not achieve the necessary performance to be true alternatives
to  manual parallelization~\cite{bhosale2022automatic}.

 The dawn of Large Language Models (LLMs), such as GPT and CodeLlama raises a fundamental question: \textbf{Can  AI-driven models revolutionize the way we approach code optimization?} These models paint an appealing horizon for code optimization and parallel code generation ~\cite{10.1007/978-981-99-8211-0_4}. Among several LLMs, we chose to explore GPT4.0 from OpenAI ~\cite{openai_chatgpt}  and CodeLlama-70B ~\cite{codeLlama_B-70} from META, as these represent the state-of-the-art and thus offer the promise of efficient code generation ~\cite{liu2023code}.

Our study compares three state-of-the-art optimizing compilers, CETUS~\cite{bae2013cetus}, PLUTO~\cite{uday08cc} and ROSE~\cite{quinlan2011rose}  against the two LLMs for parallel code optimization. We evaluate their capacity to serve as Automatic Optimization Tools (AOTs).

One of the challenges in using LLMs for code generation is the lack of correctness guarantee. To address this critical gap, we developed a mechanism for automatic validation of the generated programs. The mechanism enables LLMs to be used as AOTs and facilitates the comprehensive evaluation of these models. The mechanism can also serve as a general instrument facilitating the use of unsafe optimizations. To the best of our knowledge this is the first such  mechanism, enabling the automatic use of optimizations that do not ensure correctness.

In our evaluation, we are specifically interested in code patterns that  challenge current automatic optimization tools and in the ability of LLM-based AOTs to overcome these challenges. To this end, we have developed a benchmark suite of such patterns. Next to a comprehensive evaluation, the suite is also available to identify areas for improvement in parallel code optimization.

This paper makes the following contributions:
\begin{itemize}
    \item We present a comparative study of LLMs and traditional optimizing compilers, utilizing study cases from real-world applications, such as the NAS Parallel Benchmarks Suite (NPB) v3.3 ~\cite{bailey1991parallel}, the POLYBENCHMARK (PB) Suite v4.2 ~\cite{Polybench}, and other relevant programs.
    
   \item We introduce an automated mechanism for validating the correctness and performance of AOT-generated code called PCAOT (Performance and Correctness Evaluation of Automatic Optimization Tools).
   
   \item We present a Parallel Computing Challenge Benchmark suite (PCB) v1.0 \cite{PCBv1}, comprising 28 parallel and non-parallel use cases for evaluating the  capabilities of AOTs.
\end{itemize}

This paper is organized as follows: Section ~\ref{sec:LLMS_vs_Compilers} 
 explores  strengths and limitations of our chosen optimizing compilers and LLMs. Section ~\ref{sec:enviro_evalaution} describes our environment for evaluating automatic optimization tools and presents the Parallel Computing Challenge Benchmark suite (PCB). Section ~\ref{sec:evaluation} evaluates the  capabilities of our LLMs and the optimizing compilers on challenging code patterns for both parallel and non-parallel scenarios, followed by the discussion of related work in Section~\ref{sec:related-work}  and conclusions in Section ~\ref{sec:findings-and-conclusions}.

%% file: LLMsVsCompilers-limitations-strengths.tex
\section{LLMs vs Traditional Compilers, Strengths and Limitations }
\label{sec:LLMS_vs_Compilers}

Current optimizing compilers exhibit a mix of challenges and opportunities~\cite{monsifrot2002machine, zhao2005model,li2014feature}. Chief among them are limitations of static analysis, hampering the identification of optimization opportunities, and the complexity of deciding on best optimization patterns for  maximum program performance \cite{barakhshan2023learning}.  It is in this context that the power of LLMs emerges, offering new potential ways to approach program optimization.

Large Language Models have captured significant attention in recent years owing to their  statistical understanding of programs, offering suggestions for code generation, and facilitating software creation \cite{li2023starcoder}, \cite{li2022competition}, \cite{achiam2023gpt}, \cite{chen2021evaluating}. While LLM-generated software has demonstrated success in many applications, recent studies have highlighted the limitations in capturing crucial aspects of linguistic meaning and understanding semantic properties of the user's prompt \cite{asher2023limits}.
 
Regardless of these limitations, recent studies suggest that Large Language Models perform remarkably well in different programming tasks, such as  code generation, bug fixing, and code completion \cite{valero2023comparing}. This expertise is largely attributed to the  amount of data available on platforms such as GitHub, which provides an effective training. However, even though LLMs can generate effective code, the lack of correctness guarantees make them unsuitable for many code optimization approaches~\cite{shypula2023learning,kadosh2023advising,chen2023lm4hpc,nichols2023modeling}.


%% file: EnvironmentEvaluation.tex
\section{An Environment for Evaluating Automatic Optimization Tools}
\label{sec:enviro_evalaution}

Recall that the lack of methods for verifying the correctness of code generated by LLMs makes it difficult to compare such tools with other AOTs. Section ~\ref{sub:environment-desc} introduces a novel method and environment called PCAOT (Performance and Correctness Evaluation of Automatic Optimization Tools) that overcomes this limitation.
Additionally, we have created the Parallel Challenge Patterns Benchmark (PCB) suite v1.0 with  28 parallel and non-parallel use cases. The suite and the challenge patterns are discussed in Section ~\ref{subsec:computing-challenges}.

\subsection{Evaluation and Verification}
\label{sub:environment-desc}

Our approach entails selecting two Large Language Models (LLMs), GPT-4.0 and CodeLlama-70B, and evaluating their code optimization capabilities. The results are then compared with three different optimizing compilers, CETUS, PLUTO, and ROSE. For engaging the LLMs, we use three prompting strategies, described next.

\subsubsection{\textbf{Prompting Strategies}}
\label{sec:prompt-strategies}
We explored three different prompts associated with utilizing Large Language Models for High-Performance Computing optimization tasks. Initially, we employed a simple instruction prompt (IP) ~\cite{mishra2021reframing} ~\cite{gupta2022instructdial}. Additionally, we transformed the original simple instruction prompt into a more detailed one (DIP). Finally, we employed a Chain-of-Thought prompt  (CoT). Recent studies indicate that Chain-of-Thought prompting enhances performance across a spectrum of arithmetic, commonsense, and symbolic reasoning tasks, making it well suited for assessing domain-specific knowledge ~\cite{NEURIPS2022_9d560961}.

CETUS, PLUTO, and ROSE translate C programs to equivalent C code annotated with OpenMP parallel directives.  
Incorporating OpenMP into the prompts facilitates the comparison of the code generated by LLMs and these compilers. We chose the following prompts:

\begin{itemize}
    \item Simple Instructing Prompting (IP) - \textbf{\textit{"Given the program below, improve its performance using OpenMP."}}
    \item Detailed Instructing Prompting (DIP) - \textbf{\textit{"Given the C program below, check for read after write and write after read dependencies among iterations, if there are no dependencies among iterations of the outermost loop, parallelize this loop using OpenMP directives. If dependencies are found in the outermost loop but there exist inner loops that can be parallelized without violating data dependencies, then parallelize those inner loops instead."}}
    \item Chain-of-Thought Prompt - \textbf{\textit{"<DIP> + As you work through the program, explain each step of your reasoning process to ensure clarity and correctness in your optimization decisions.  Think step by step."} }
\end{itemize}

The main difference between the three types of prompts recline in their ability to guide the model’s reasoning process. Instructing prompts provide a more direct and clear directive, resulting in the immediate production of an optimized program version. As opposed to Chain of Thought (CoT) prompts, which enable the model to break down the task into multiple steps, allowing for iterative refinement before generating an optimized program version. Instances of guidance in the model in this process are: \textbf{"Explain each step of your reasoning process...", "Break down your thought process into individual steps...", "Detail the reasoning behind your answer..."} ~\cite{shypula2023learning}.

We chose to add a condition into the prompts DIP and CoT which is \textbf{"check for read after write, and write after read dependencies among iterations, If there are no dependencies among iterations of the outermost loop, parallelize this loop using OpenMP directives...."}. This guidance consists of two key components: first, analyzing data dependencies to ensure correctness, and second, prioritizing the parallelization of outer loops.  These key components represent two of the most fundamental concepts in automatic program parallelization. Additionally, when prompted simply to \textbf{"improve its performance"}, with IP, LLMs are likely to attempt parallelization regardless of whether the code is actually parallelizable. The DIP and CoT prompts express high level compiler strategies that guide the LLMs to improve their reasoning process. 



\begin{figure*}
\includegraphics[width=15.5cm,height=4.0cm]{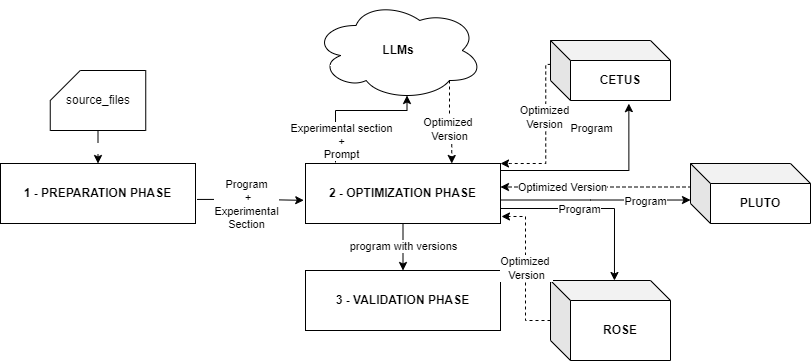}
\centering
\caption{Architecture of the PCAOT Environment.}
\label{fig:mesh3}
\end{figure*}

For the optimizing compilers, we used the default set of parameters, representing the known, safe optimization methods. Note that, while experimenting with optimization options could potentially yield improved performance, doing so is beyond the scope of this study.
Our environment for evaluating the five AOTs comprises three phases: \textbf{Preparation,  Optimization}, and  \textbf{Validation}. Figure~\ref{fig:mesh3} illustrates this architecture.

\subsubsection{\textbf{Preparation Phase}}

LLMs have shown to have limitations in processing programs with large code sizes ~\cite{openaiTokens} ~\cite{codeLlama_70b}. Hence, our LLM evaluation focuses on section-level optimizations.
The preparation phase identifies the sections within the source code to experiment with.  Each such section is manually enclosed with \textbf{\#pragma experimental section start} and ~\textbf{\#pragma experimental section stop}, as illustrated in Figure~\ref{fig:fig1}. The impact of code size on the performance of the LLMs will be discussed in Section~\ref{sec:codesize-impact}.

\begin{figure}[H]
    \centering
    \footnotesize
    \begin{verbatim}
  1: #pragma experimental start 
  2: for (k=0; k<=(grid_points[2]-1); k ++ )
  3:  {
  .      ...computation
  9:  }
  10: #pragma experimental end 
    \end{verbatim}
    \vspace{-5mm}
    \caption{Experimental section instrumentation. {\normalfont Example taken from subroutine PO\_Function\_Version\_1 of the PCB suite.}}
    \label{fig:fig1}
\end{figure}

 For our experiments, we made use of a tool called CaRV (Capture, Replay, and Validate), which enables users to experiment with sections of large applications. It facilitates the comparison of individual program sections before and after optimizations, assessing their efficiency and accuracy ~\cite{barakhshan2023carv}. CaRV does so using a selective checkpointing method, based on the identified program states at the beginning and end of an experimental section. The preparation phase generates a new program version that is instrumented in a way that allows the optimized program section to be run independently and compared against the captured end state of the original program. Both the original and instrumented program versions are passed on to the optimization phase.


 
\subsubsection{\textbf{Optimization Phase}}

 This phase sends the experimental section with the prompts to the two LLMs to obtain the optimized versions. The resulting code
 is then inserted into the instrumented program.  The original code is also sent to the three compilers, CETUS, PLUTO, and ROSE. This process generates a total of 22 program versions, for the five AOTs (two LLMS, three different prompts, three requests per prompt, three optimizing compilers plus the serial program).
 The next phase executes and validates these versions.

\subsubsection{\textbf{Validation Phase}}


The Validation phase executes each version of the experimental section individually. The LLM-generated versions are then compared with the original, serial program versions, using the CaRV checkpoints. This process verifies the values of all relevant, "live" variables at the end of the LLM-optimized program section against the output state of the corresponding values in the original program.  The execution time is measured as well and compared to that of the original program section.

\subsection{Parallel Computing Challenge Pattern Benchmark Suite v1.0 (PCB)}
\label{subsec:computing-challenges}
The PCB Suite was specifically created to address a critical need for unbiased evaluation by ensuring that it has not been exposed to LLMs during training. This new Benchmark suite comprises 28 distinct use cases, each presenting one of six challenging scenarios encountered by optimizing compilers ~\cite{barakhshan2023learning}. A number of patterns are included in the PCB suite, testing the optimizers in their ability to: \textbf{parallelize outermost loops (PO), parallelize loops with function calls (PF), form parallel regions enclosing multiple parallel loops (PR), parallelizing array reductions (PA), avoiding load imbalance through dynamic scheduling (DS),} and \textbf{eliminate barrier synchronizations using NOWAIT clause (NW)}.


These patterns were identified as key challenge areas for classical optimizing compilers ~\cite{barakhshan2023learning}, ~\cite{prema2017identifying}, an important question is: \textbf{ Can LLMs identify these patterns and perform the requisite optimizations?}
More detailed explanation of these challenge patterns can be found in prior studies \cite{barakhshan2023learning}.

%% file: LLMsVSCompilers.tex
\section{Evaluation: LLMs versus Classical Optimizing Compilers}
\label{sec:evaluation}
This section assesses the current capabilities of two Large Language Models, GPT-4.0 and CodeLlama-70B, in comparison with three different automatic compilers, CETUS, PLUTO and ROSE. Section ~\ref{sec:exp-setup} outlines the experimental setup, Section ~\ref{sec:codesize-impact} evaluates the ability of the LLMs to process and correctly optimize code of different sizes. Section ~\ref{sec:correctness-evaluation} examines the ability of these LLMs to optimize code sections effectively and also assesses their capacity in handling the challenge patterns introduced in  Section~\ref{subsec:computing-challenges}. Lastly, Section ~\ref{sec:speedup-evaluation} compares the achieved speedup of all five Automatic Optimization Tools (AOTs).


\subsection{Experimental setup}
\label{sec:exp-setup}

To assess our Automatic Optimization Tools, we use the PCAOT environment described in Section ~\ref{sub:environment-desc}. Our test suite drew from three distinct benchmarks: The NPB v3.3~\cite{bailey1991parallel} , the PB suite V4.2 ~\cite{Polybench}, and the PCB benchmark introduced in Section~\ref{subsec:computing-challenges}. We measured the performance of the applications using input Class B for the NPB and the LARGE\_DATASET for the PB and PCB suite. 
Recall that we are focusing on challenge patterns for the optimizing compilers. We have selected specific subroutines from the NPB and the PB suites that contains such cases with a total of 16 experimental sections. For the PCB, we developed 28 use cases, with each one representing one of the six challenge scenarios described in Section~\ref{subsec:computing-challenges}. In total, our testing dataset contains 44 experimental sections. More details about the testing data set can be found on our GitHub repository \cite{testingDataSet}.


We set the following AOT parameters: For the CETUS, PLUTO and ROSE compilers we chose their default options, selecting the known, safe optimization behavior.  For the LLMs, we chose a Temperature of \textbf{0.2} and a Top\_p value of \textbf{0.1}, increasing deterministic behavior for the code generation tasks.

We measured execution times using 4 cores on a compute node featuring an Intel Xeon Gold 6230 processor configuration in dual sockets. Each processor operates at a base frequency of 2.1 GHz, with a 27.5MB cache, and was supported by up to 1 GB of DDR4 memory. Application codes were compiled using GCC v11.2 with -O3 optimization on CentOS v7.4.1708. Reported values represent the median of three application runs, each utilizing one thread per core.

\subsection{Impact of Code Size}
\label{sec:codesize-impact}

Figure \ref{fig:codeSizesFig}, shows the failure rate of the Large Language Models in generating programs with increasing code sizes. An unsuccessful case refers to instances where the LLMs failed to produce output that passed validation. The graph displays the number of lines of code per experimental section versus the unsuccessful cases. We sent the sections to GPT4.0 and Codellama-70B, using the three different prompts described in Section ~\ref{sec:prompt-strategies}, repeating each run three times, with a total of 792 attempts (44 experimental sections x 3 times each x 3 prompts x 2 LLMs = 792 attempts). Details about the experimental sections used in this study can be found on our GitHub repository \cite{testingDataSet}.

These results picture the LLMs' capability to handle large code sizes. While smaller code sections tend to be transformed more accurately, they are not guaranteed to be 100\% correct. As the code size increases, the likelihood of inaccuracies also increases.

Program sections of approximately 20 lines often produce correct output, whereas sections above 20 lines increase the failing percentage (performance will be evaluated later). In between, the LLMs exhibit non-deterministic behavior, with some runs producing useful results while others with the same input fail.
GPT4.0 and Codellama-70B return successfully transformed code
in 16.75\% of the cases. These results corroborate our method of employing LLM capabilities on the basis of individual program sections, such as loops.

\input{plots/sizesKDE}

\subsection{LLM Optimization Capabilities}
\label{sec:correctness-evaluation}

In this subsection, we evaluate the capacity of Large Language Models (LLMs) to optimize both parallel and non-parallel scenarios. Parallel use cases represent scenarios where the LLMs must identify opportunities for parallelization and apply the correct optimization strategy. In contrast, non-parallel use cases test the LLMs capability to accurately recognize scenarios where parallelization is not feasible or beneficial. For this purpose, we utilized 94 parallel use cases and 10 non-parallel use cases. 


\subsubsection{\textbf{Parallel Use cases:}}
\label{subsec:positive_usecases}
Figure \ref{fig:CorrectnestEv} shows the success rates of generating correctly optimized code by the tested LLMs on specific challenge patterns, using the three prompts described in Section~\ref{sec:prompt-strategies}. The results are classified into three distinct categories: \textbf{Expected Pattern applied}, \textbf{  Unexpected Pattern Correctly Applied} and \textbf{Errors} (Compilation and Runtime Errors).

\input{plots/errors_plot}

Our primary finding from this information is that LLMs are not yet in a position where they can serve as automatic optimizers. Only in three of the analyzed optimization patterns, both LLMs produce partly correct results. A key demand on automatic optimizers is that they be 100\% correct; tools that require software engineers to engage in possibly lengthy debug phases after use are not acceptable. The partial successes in three of the patterns (Parallelizing Outermost Loop, Parallelization with Function Calls and Array Reduction) indicate that LLMs may be making progress toward the goal of becoming useful AOTs. These three patterns represent some of the most important parallelization capabilities and likely have many code examples serving as LLM training data, which may explain the partial success. The Parallel-Region and NOWAIT patterns 
tend to involve large code regions, which challenge LLMs, as discussed in Section~\ref{sec:codesize-impact}. The two LLMs completely fail on Dynamic Scheduling patterns. We attribute this behavior to the more complex code exhibited by such test cases. Dynamic scheduling tends to be needed in loops that involve irregular code and data structures, which are more difficult to comprehend.

The results show a substantial difference between the three prompting strategies, with DIP being better among the three different approaches in most of the cases. While other researchers have reported more success with CoT prompting, this method performed essentially similarly to DIP. 


\subsubsection{\textbf{Non-parallel Use cases}}

Figure~\ref{fig:negative_usecases} illustrates the correctness of the non-parallel use cases. For this analysis, we evaluate loops with data dependencies (DP) and loops with non-side effect free function calls (FC).


\begin{itemize}
    \item \textbf{Non\_Parallel\_Loop\_DP:} This category represents code sections with data dependencies that prevent safe parallelization.
    \item \textbf{Non\_Parallel\_Loop\_FC:} This category represents code sections containing function calls that are not side-effect-free, thereby making data dependence analysis required for safe parallelization more complex.

\end{itemize}

These categories help differentiate the specific reasons why loops in the negative use cases cannot be parallelized, assessing the LLMs' ability to correctly recognize non-parallelizable scenarios.
The results show that the DIP strategy performed marginally better than the CoT approach in recognizing non-parallel use cases, a similar result was found for the parallel use cases show in Section ~\ref{subsec:positive_usecases}. However, both LLMs failed entirely to identify non-parallel code when using IP. We attribute this to the lack of specificity in the IP prompt, as explained in Section ~\ref{sec:prompt-strategies}. 


\input{plots/negative}

\subsection{\textbf{Speedup Evaluation by Pattern}}
\label{sec:speedup-evaluation}

Figure \ref{fig:speedup_patterns} shows the maximum average speedups of the correctly optimized codes for the parallel scenarios from the two chosen LLMs. These scenarios include GPT4.0 and CodeLlama-70B across our testing Dataset described in \cite{testingDataSet} with the three different prompts described in Section \ref{sec:prompt-strategies} for three specific patterns: Parallelization at the Outermost level, Array Reduction and Parallelization for function calls.

\input{plots/speedup_patterns}

Our results demonstrate the efficacy of CodeLlama-70B, which achieves speedups of up to \textbf{1.95x} and \textbf{1.79x} over the original program. Additionally, the DIP prompt is more efficient for this particular AOT. We attribute this behavior to the type of model that is being used. CodeLlama-70B is fine-tuned to follow instructions more closely.  \cite{codeLlama_70b}. 

\subsection{\textbf{Speedup Evaluation for each AOT}}

Figure \ref{fig:speedup-general} shows
the maximum average speed-up of the correctly Parallel optimized codes  obtained from the five distinct AOTs over the serial and the Hand-optimized versions.   

Our results show CodeLlama-70B as the superior optimizer among Language Model-based (LLM) solutions, capable of achieving speedups of up to \textbf{1.75x} compared to the original program.

Conversely, we can observe a noticeable difference between all AOTs and PLUTO in terms of the speedup achieved within the PB suite, where PLUTO outperformed all AOTs with a remarkable \textbf{7x} speedup. This superior performance is attributed to PLUTO's specialization on the Polyhedral model~\cite{10.1007/978-3-540-78791-4_9}. The specialization may also explain why PLUTO was unable to process the other benchmark suites.

Overall, CETUS emerged as the best among the optimizing compilers for processing a variety of programs within these suites, achieving a maximum speedup of \textbf{1.67x}. The ROSE optimizer achieved a speedup across all benchmarks of \textbf{1.17x}, remaining below the other AOTs in terms of efficacy. 

While GPT4 has demonstrated success across various tasks, it exhibited the lowest performance for code optimization across all the benchmarks and among the LLMs. It achieved a maximum speedup of \textbf{1.34x}, falling short of CodeLlama-70B.

\input{plots/speedup_plot}

%% file: plots/sizesKDE.tex
\begin{figure*}[htbp]
     \caption{Average failure rate of the Large Language Models across benchmarks when generating programs with increasing code sizes.}
    \label{fig:codeSizesFig}
    \includegraphics[width=8cm,height=5cm]{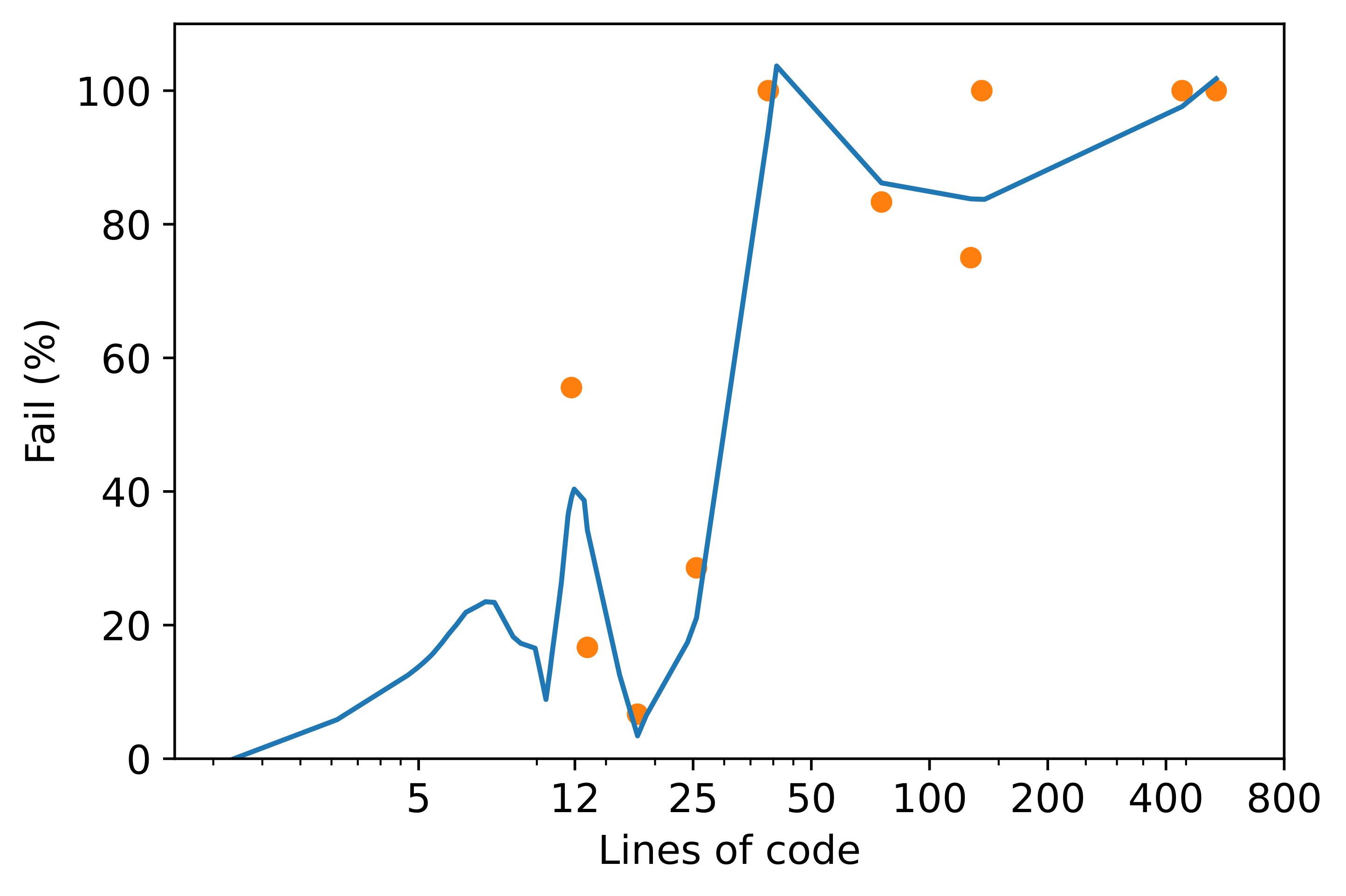}
     
\end{figure*}

%% file: plots/errors_plot.tex
\begin{figure*}[htbp]
    \centering
    \caption{Correctness Evaluation for the challenge patterns: Parallelization at the Outermost level (P-Outermost), Parallelization for function calls (P-Function Calls), Array Reduction,  Parallel-Regions, NOWAIT, and Dynamic Scheduling (Dynamic Sch).}
    \label{fig:CorrectnestEv}
    \includegraphics[width=13cm,height=7cm]{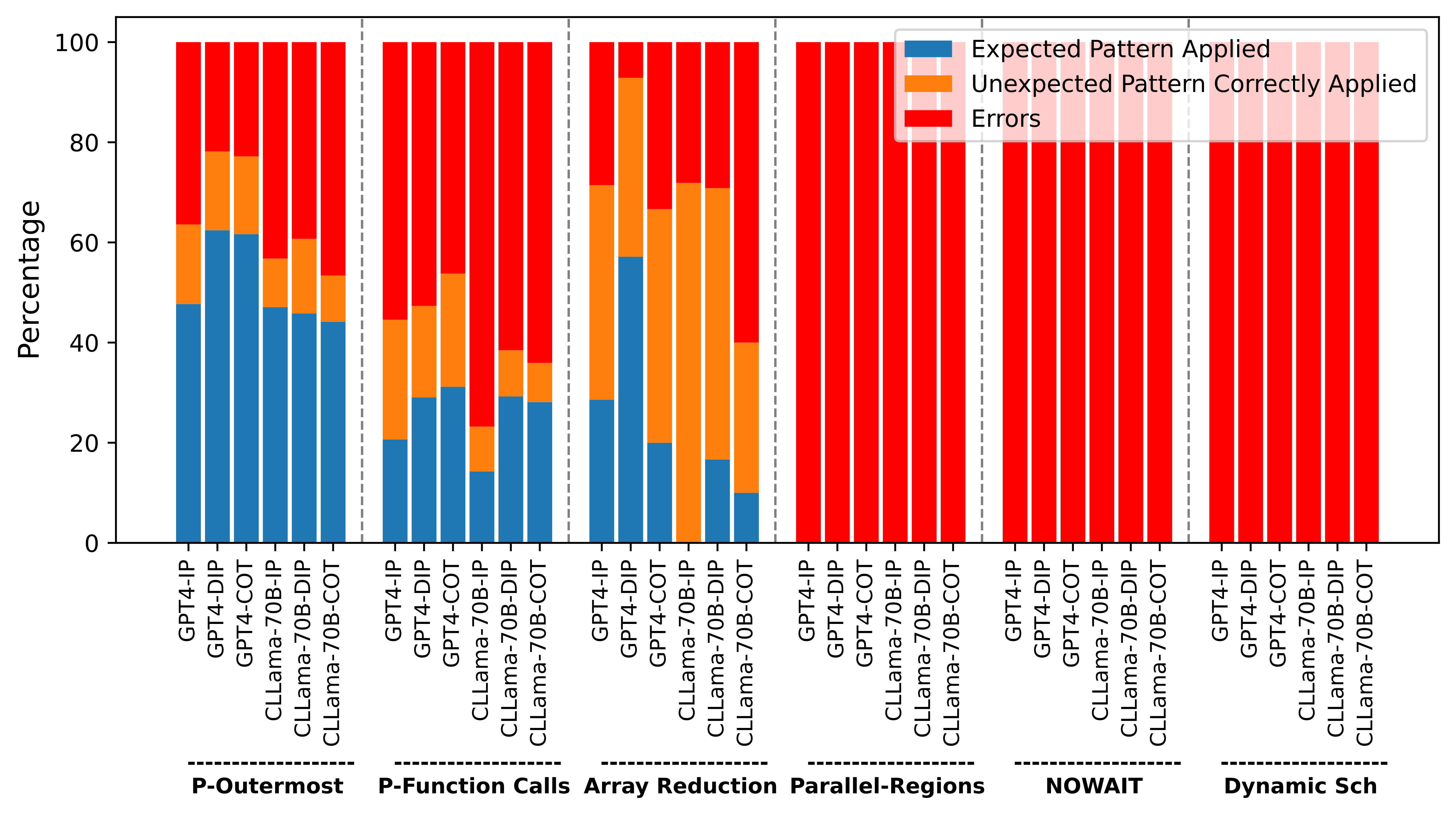}

\end{figure*}

%% file: plots/negative.tex
\begin{figure*}[htbp]
    \centering
    \caption{Correctness Evaluation for the  challenge patterns: Parallelization at the Outermost level and Parallelization for function calls in non-parallel Scenarios. The expected behavior in this case is that the LLM recognizes the code as non parallelizable. }
    \label{fig:negative_usecases}
    \includegraphics[width=10cm,height=5cm]{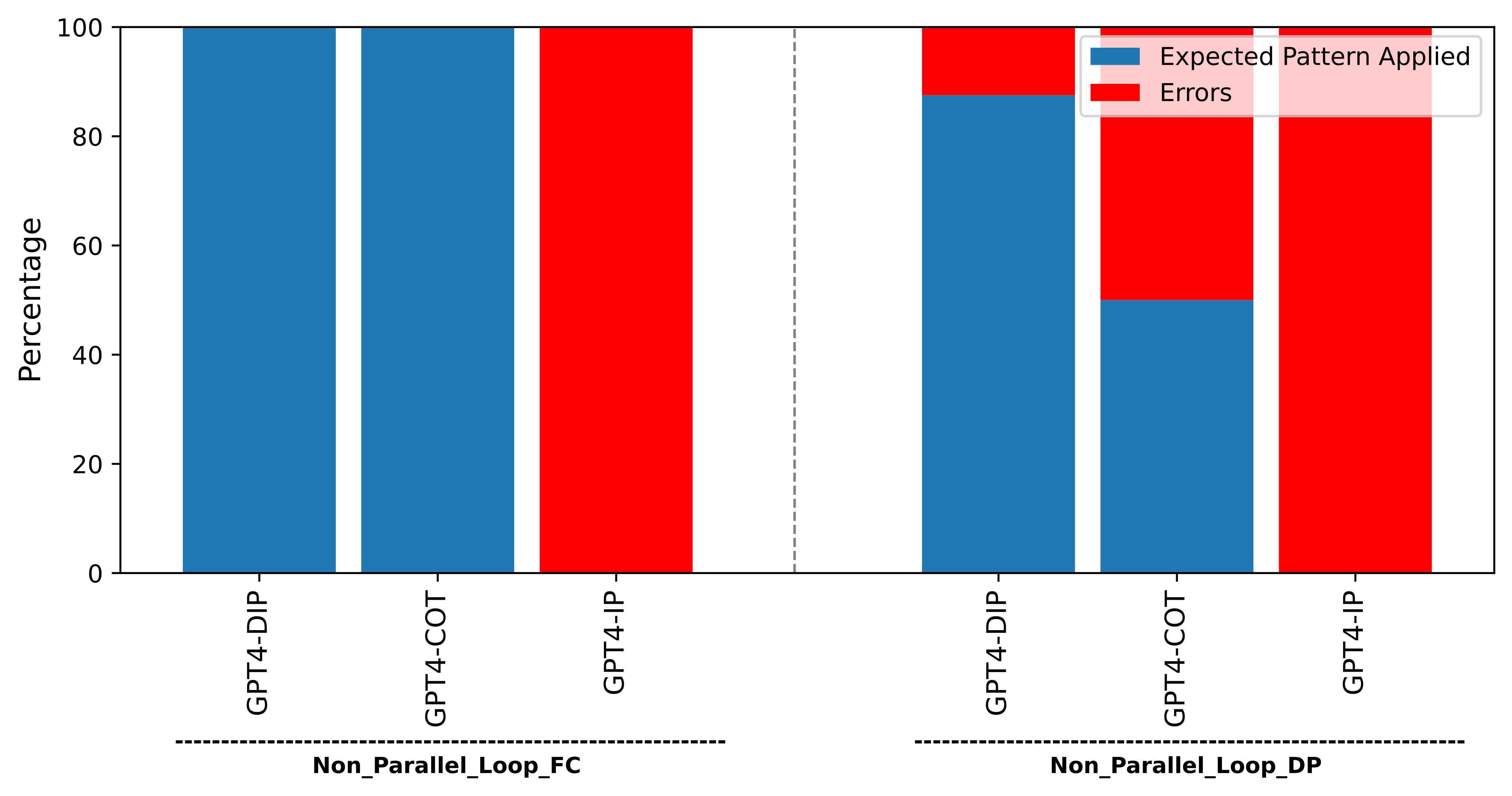}
\end{figure*}

%% file: plots/speedup_patterns.tex
\begin{figure*}[htbp]
    \centering
    \caption{Speedup Evaluation two LLMs (AOTs) across three challenging patterns}
    \label{fig:speedup_patterns}
    \includegraphics[width=14cm,height=4cm]{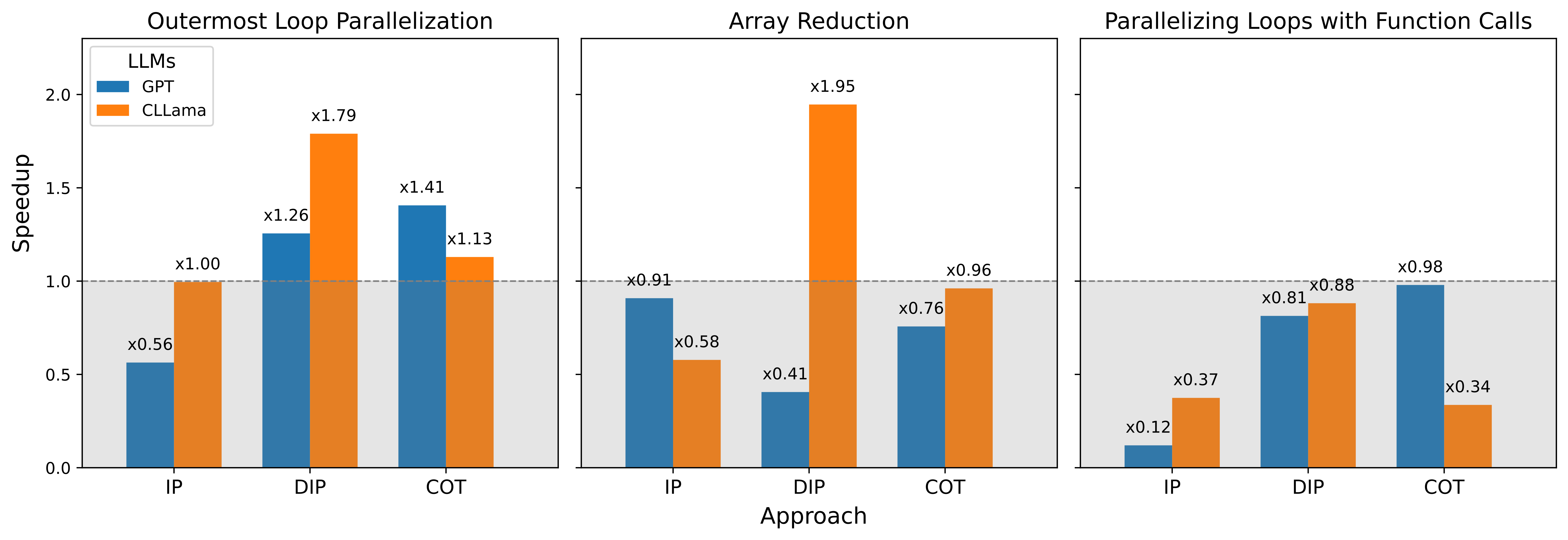}

\end{figure*}

%% file: plots/speedup_plot.tex
\begin{figure*}[htbp]
    \centering
    \caption{AOTs - Speedup Evaluation of each AOT using a specific configuration: Simple Instruction, Instruction prompting, and Chain-of-Thought for LLMs and the default configuration for the optimizing compilers.}
    \label{fig:speedup-general}
    \includegraphics[width=8cm,height=5cm]{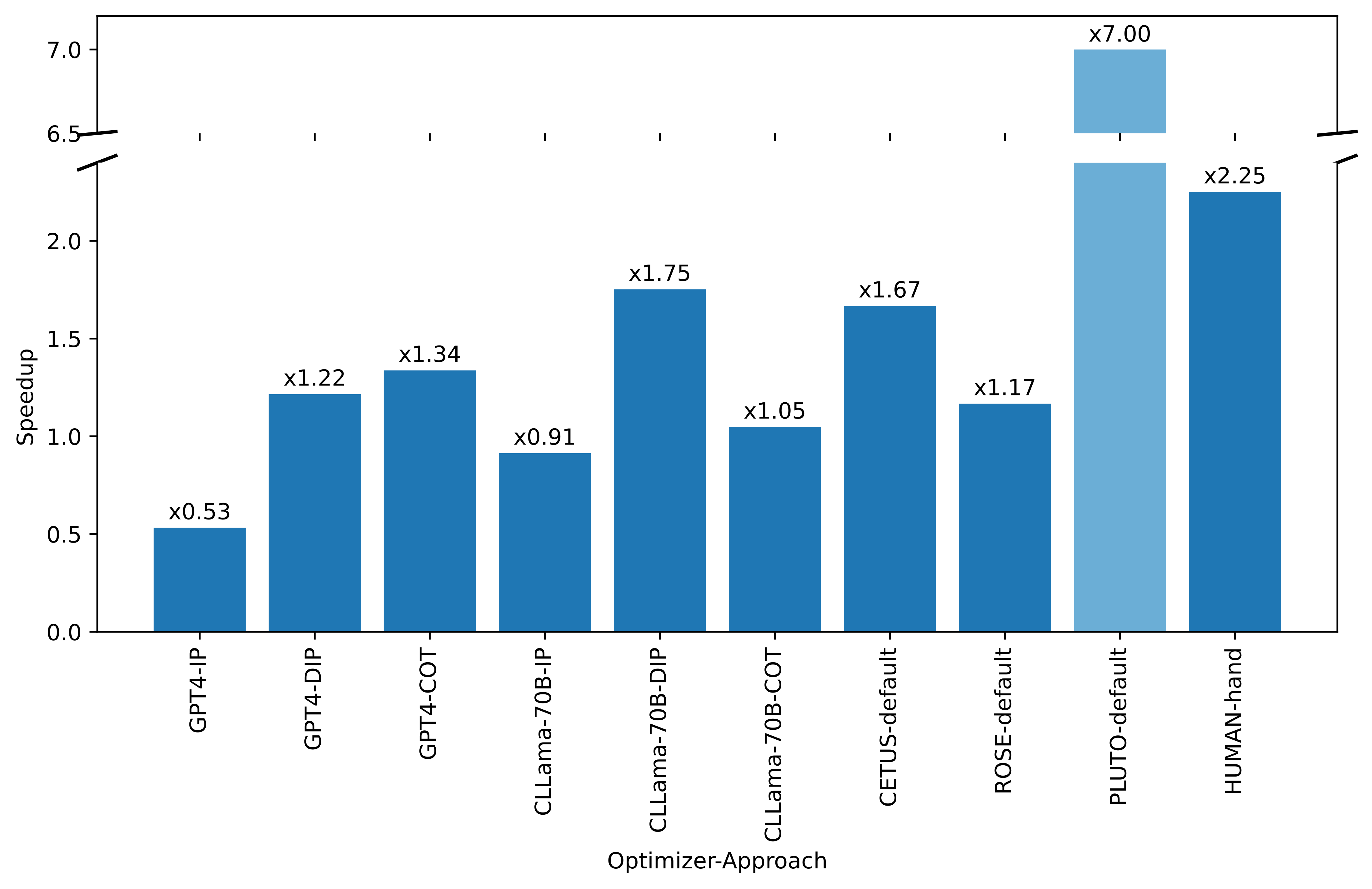}

\end{figure*}

%% file: Related_Work.tex
\section{Related work}
\label{sec:related-work}



Recent studies have evaluated the performance of LLMs in specific High-Performance Computing (HPC) optimization tasks, such as OpenMP pragma prediction \cite{nichols2023modeling}  highlighting the application of LLMs in parallel computing optimizations. Additionally,  approaches to code completion \cite{valero2023comparing} and issue fixing \cite{valero2023comparing} show the breadth of LLM applications in software development and maintenance. 

Another significant contribution in this domain is the novel paradigm introduced by researchers using LLMs with compiler feedback to optimize the code size of LLVM assembly. Their model takes unoptimized LLVM Intermediate Representation (IR) as input and produces optimized IR, optimization passes, and instruction counts for both unoptimized and optimized IRs ~\cite{grubisic2024compiler}. 

Recent studies have evaluated the ability of LLMs to generate short programs from natural language descriptions, demonstrating the potential of these models in code generation tasks \cite{austin2021program}. However, despide of these advancements, the automatic validation of both the correctness and performance of optimizations remains underexplored.

The primary objective of PCAOT is to automatically validate the correctness and performance of optimizations generated by Automatic Optimization Tools (AOTs). Our work aims to fill this gap by providing a comprehensive framework for evaluating and validating the outputs of AOTs, ensuring that the generated optimizations meet the desired standards of correctness. To the best of our knowledge, no other research
has directly addressed this specific objective. 

Last but not least, several benchmarks have been proposed to evaluate the capacity of AOTs on their abilities and limitations in optimizing code for maximum efficiency ~\cite{bailey1991parallel}, ~\cite{Polybench}, \cite{black2022gpt}. We  proposed a novel Parallel Computing Challenge Benchmark suite (PCB) V1.0 containing 28 use cases, each representing one of the six challenge patterns for evaluating AOT code optimization capabilities. To the best of our knowledge, no other benchmark have been designed to directly addressed this specific goal.



%% file: finding-and-conclusions.tex
\section{Conclusions}

\label{sec:findings-and-conclusions}

This paper addresses an existential question of the compiler community: \textit{"Can AI-driven models revolutionize
the way we approach code optimization?"} To this end we evaluated five different Automatic Optimization Tools (AOT - three classical compilers and two LLMs), assessing their respective abilities and limitations
in optimizing code for maximum efficiency. To support this evaluation, we introduced a novel mechanism called PCAOT, which enables the automatic validation of the correctness and performance of code generated by AOTs. While we have applied the mechanism to verify LLM-generated code, it also enables unsafe compiler optimization to be tested. Additionally, we have presented the Parallel Computing Challenge Benchmark suite (PCB) v1.0, comprising 28 parallel and non-parallel use cases for evaluating the capabilities of AOTs. To the best of our knowledge, this is the first such benchmark.

Our results show that LLMs only succeed in correctly optimizing very small programs of approximately 20 lines.
Given that real-world software often contains thousands of lines of code, 
these models are not yet ready to serve as automatic optimizers. 

 In terms of performance, the chosen LLMs can successfully identify three out of the six different challenge patterns, including Array Reduction, Parallelizing Outermost Loops, and Loops Containing Function Calls, with a success rate higher than 30\% in most cases. However, they completely fail on other patterns, including eliminating barrier synchronization using the NOWAIT clause, parallel regions enclosing multiple parallel loops, and avoiding load imbalance through dynamic scheduling. 

We found substantial differences between three prompting strategies we employed for both the parallel and non-parallel scenarios, with "Detailed Instruction Prompting" (DIP) being better
among the three different approaches in most of the cases. Additionally, "Chain-of-Thought" (CoT) prompting performed essentially similarly to DIP. We attribute this behavior to the high-level compiler knowledge strategies expressed in these two prompts.

In terms of speedup,  the best compiler and the best LLM performed within 10\%. CodeLlama, which is specifically optimized for code synthesis and infilling/completion tasks~\cite{grattafiori2023code}, was expected to demonstrate strong performance in this context. However, ChatGPT contains a larger number of AI model parameters, which may make up for the fact that it is a general model.


Finally, addressing the main question asked in the introduction, we find that, with their current capabilities, the LLMs cannot yet suitably replace classical compiler optimization technology. Even though the potential exists, future LLMs will need significant improvements in terms of both performance and correctness. Future work will explore additional factors, such as the energy consumption required to achieve performance gains over conventional optimizing compilers. We also anticipate that our results will remain consistent across different platforms.

%% file: acknowdledgements.tex
\begin{acks}
\label{sec:acknowdledgements}

This work was supported in part by the University of Delaware and by the National Science Foundation under awards 2112606, 2125703, and 1931339.

\end{acks}